\documentclass[letterpaper, 10 pt, conference]{ieeeconf}  

\IEEEoverridecommandlockouts                              

\usepackage{amsmath} 
\usepackage{amssymb}  
\usepackage{bm}  
\usepackage{graphicx}
\usepackage{multirow}
\usepackage{url}
\usepackage{subcaption}
 \usepackage{cite}
\usepackage{lipsum}

\DeclareMathOperator*{\argmax}{arg\,max}

\title{\LARGE \bf
Multi-Objective Compliance-Integrated Coevolution For Simulated And Real-World Deployment Of Multi-Robot Marine Autonomy
}

\author{Everardo Gonzalez$^{1}$, Tyler M. Paine$^{2}$, Manuel Agraz Vallejo$^{1}$, \\Gaurav Dixit$^{1}$, Michael R. Benjamin$^{2}$, and Kagan Tumer$^{1}$
\thanks{$^{1}$Everardo Gonzalez, Manuel Agraz Vallejo, Gaurav Dixit, and Kagan Tumer are with Collaborative Robotics And Intelligent Systems Institute, Oregon State University, Corvallis, OR, USA 
{\tt\small \{gonzaeve,agrazvam,dixitg,kagan.tumer\}}
{\tt\small @oregonstate.edu}
}%
\thanks{$^{2}$Tyler M. Paine, Michael R. Benjamin are with MIT Marine Autonomy Lab, MIT, Cambridge, MA, USA {\tt\small \{tpaine, mikerb\}@mit.edu}}%
}

\usepackage{algpseudocode}
\usepackage[ruled,vlined,linesnumbered]{algorithm2e}

\begin{document}

\maketitle
\thispagestyle{empty}
\pagestyle{empty}

\begin{abstract}
Collaborative robots are well-suited to maritime missions that benefit from coordination, such as the exploration of unknown reef structures, inspection of subsea infrastructure, or search-and-rescue operations.
These missions typically provide sparse feedback signals for measuring progress and require adherence to safety and regulatory norms, turning a mission into a multi-objective optimization problem. 
Coevolutionary algorithms can process these sparse feedback signals to generate coordinated behaviors, and in some cases extend behaviors to multiple objectives.
However, incorporating high-level team objectives with low-level compliance considerations on the fly to balance norm adherence with team performance remains elusive.
This paper introduces a multi-objective framework that blends coevolved behaviors with compliance behaviors to achieve a balance between maximizing team progress and minimizing norm violations.
The key insight is to decouple learning from compliance since operational norms are prescribed rather than discovered.
We demonstrate that our framework achieves high team performance while avoiding collisions on a collaborative swimmer rescue mission with up to 8 vehicles in a hardware deployment, and 12 vehicles in simulation.
The key contribution of this paper is Marine Multi-Objective Compliance-Integrated Coevolution (MMOCIC), a framework that blends team-wide optimization with established norms for real-world deployments of learning-based coordination.
\end{abstract}

\section{INTRODUCTION}

Collaborative robots are well suited for maritime missions that require coordinated operation across large areas, including exploration of unknown environments, inspections of subsea infrastructure, and search and rescue.
Many such missions are characterized by a sparsely defined team objective (``sparse'' meaning success is measured through delayed and uninformative feedback) and explicitly defined operational norms. 
Each robot must balance this sparse team objective with compliance directives to make informed tradeoffs between team performance and norm adherence. 
For instance, we may need a rescue team to comply with COLREGs (maritime right-of-way regulations \cite{USCG_NavRules_Handbook_2024}), but also be able to perform risky maneuvers that intentionally violate COLREGs if necessary to rescue a drowning swimmer.

Cooperative coevolutionary algorithms partially address this challenge by optimizing individual robot behaviors using a sparsely defined team objective \cite{agogino2004gecco, potter200evo, kelly2022gecco}.
Unlike reinforcement learning methods that depend on carefully shaped, dense reward signals \cite{gupta2018oceans, wang2021neuralnet, zhou2019access, zheng2025iros, chu2025iros}, cooperative coevolution operates directly on sparse mission objectives. 
However, it does not inherently provide a mechanism for incorporating compliance into real-world deployments, which turn a mission into a multi-objective optimization problem \cite{hu2017ifac}.

\begin{figure}[t]
    \centering
    \includegraphics[width=0.9\linewidth]{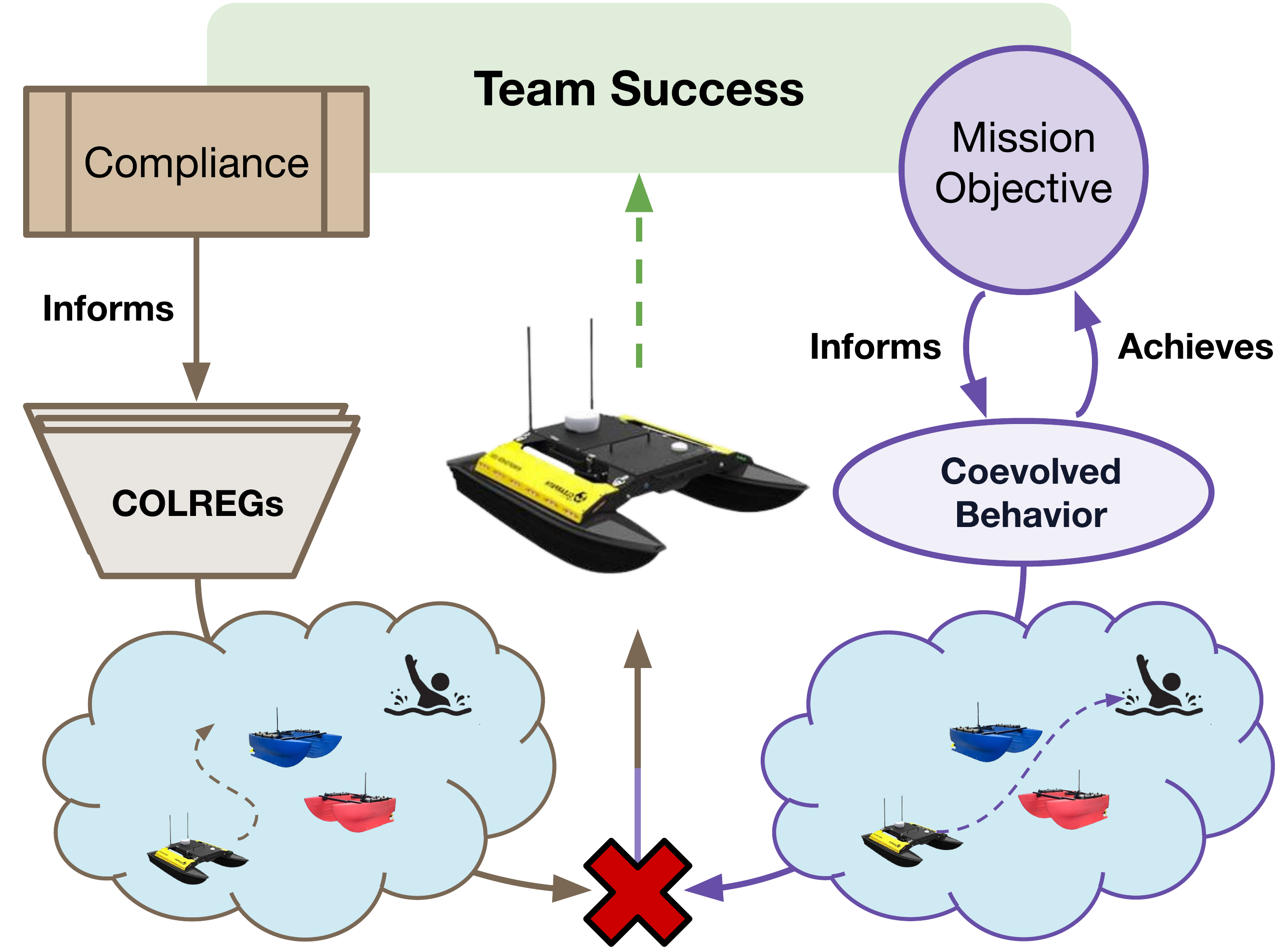}
        \caption{\textbf{Problem Setting.} Team success depends on each vehicle (yellow Heron) simultaneously rescuing swimmers and adhering to maritime regulations. 
        A mission objective drives a coevolved behavior to efficiently navigate toward swimmers, potentially requiring maneuvering through surrounding traffic. Compliance requirements, informed by COLREGs, prescribe interactions that govern safe passage relative to nearby vessels (red and blue boats). These two perspectives, mission achievement and regulatory adherence, are distinct and exert competing pressures on each vehicle.
        }
    \label{fig:problem-illustration}
\end{figure}

In this work, we introduce Multi-Objective Compliance Integrated Coevolution (MMOCIC), a behavior mixing framework that smoothly transitions marine robots between coevolved collaborative behaviors and prescriptive compliance-focused behaviors.
The core mechanism blends behaviors through a continuous weighting function to achieve a balance between achieving a mission objective, minimizing the risk of collisions through adherence to COLREGs, and staying within a specified operating region.
In hardware deployments, the output of this weighting function is filtered through a collision-avoidance controller that overrides control input if a collision is imminent.

The key insight is to decouple learning from compliance so we can evolve expressive collaborative behaviors in simulation without constraining their scope, while ensuring deployed robots adhere to expected norms. 
Since operational norms are prescribed, there is no need to try to discover them through coevolution. Instead, we delegate the challenging coordination discovery problem to coevolutionary optimization, and norm adherence to other compliance focused behaviors.

We evaluate MMOCIC on a collaborative swimmer rescue mission illustrated in Figure \ref{fig:problem-illustration}. We show two sets of experiments on hardware to demonstrate feasibility of real world deployment: 1) 8-robot team in a tight operating region, and 2) 4-robot team with a disruptive agent. 
We further present a zero-shot ablation study with 12 simulated robots to show how weight adjustments enable tunable tradeoffs between a mission objective and compliance without re-training.

\section{RELATED WORK}

\subsection{Optimizing Sparse Objectives}

Evolutionary algorithms optimize behaviors using episodic fitness evaluations, meaning they search directly over the behavior space to associate behaviors with outcome-based objectives that provide only sparse feedback \cite{kelly2022gecco, zhu2021gecco, shao2025acm}.
Sparse feedback is delayed and infrequent, often arriving only after an extended sequence of joint actions \cite{zhu2021gecco, rahmattalabi2019iros}.
Reinforcement learning struggles in these settings because it typically requires dense reward structures to associate each action an agent takes with immediate feedback \cite{gupta2018oceans, wang2021neuralnet, zhou2019access, zheng2025iros, chu2025iros}.

Cooperative Coevolutionary Algorithms (CCEAs) \cite{agogino2004gecco, potter200evo, kelly2022gecco} decompose team optimization into agent-level subproblems. Behaviors are evolved in separate populations and jointly evaluated. Fitness shaping techniques further amplify gains by modifying individual agent fitnesses according to their estimated contribution to the team \cite{agogino2004gecco, rahmattalabi2019iros}.
Additionally, Evolutionary Multiobjective Optimization can approximate a Pareto front of solutions that represent tradeoffs for decision support, where we achieve higher performance on one objective only at the cost of degrading another objective \cite{kesireddy2019ssci, du2023complex, shao2025acm, hu2017ifac}.
Unfortunately, compliance is formulated as a set of low-level instructions rather than high-level objectives, so they cannot be ingested by this type of optimization.

\subsection{Multi-USV Autonomy}

Learning-based approaches for Unmanned Surface Vehicles (USVs) often focus on tasks such as target tracking or capture the flag, where reward functions are extensively shaped around distance, coverage, or tracking metrics \cite{wang2022icus, zhu2024rcar, gong2025ieee, beason2024arxiv, gupta2018oceans}.  
Learned behaviors are tightly coupled to carefully tuned low-level reward functions rather than long-term team outcomes.
They are difficult to adapt without re-training, and struggle to incorporate sparse objectives or compliance.

Multi-USV deployments typically balance multiple objectives through behavior based autonomy \cite{benjamin2010fieldrobotics, kim2017ut, beason2024arxiv}.
Behavior based autonomy formulates each behavior as a dense local objective that gets combined into one objective that a vehicle can locally optimize.
This makes it possible to incorporate low-level compliance requirements directly into each USV \cite{colregsbehavior, opregionbehavior}.
Control Barrier Functions (CBFs) are similarly promising because they add a layer of theoretical safety guarantees in many multi-USV coordination problems \cite{gao2022automatica, ding2024icra}.
However, neither method provides a way of optimizing long-term mission outcomes because behaviors must be specified apriori and CBFs are limited to constraints.

\begin{figure*}[ht!]
    \centering
    \includegraphics[width=1.0\linewidth]{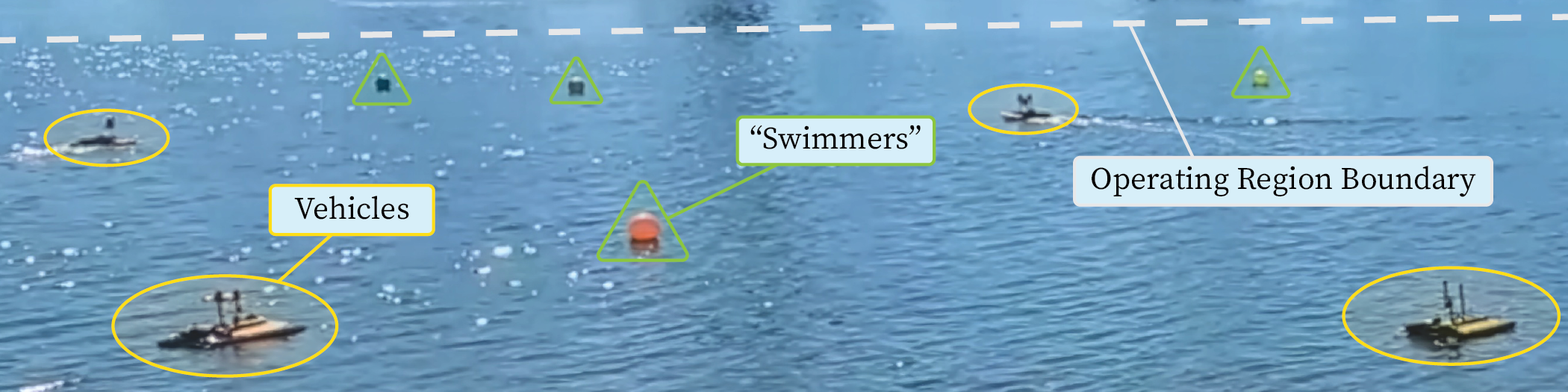}
        \caption{\textbf{Collaborative swimmer rescue mission.} Buoys are placed at the locations of swimmers (marked with green triangles). A team of autonomous vehicles (marked with yellow ovals) must coordinate to rescue randomly placed swimmers in a fixed time. Buoys were removed for experiments, but help provide an intuitive sense of scale.}
    \label{fig:real-world-overview}
    \vspace{-2mm}
\end{figure*}

\section{PRELIMINARIES}

\subsection{Behavior-Based Autonomy}
In the context of layered behavior-based autonomy \cite{brooks1986robotics}, objective functions express preferences by mapping the set of available actions to a scalar utility. An objective in behavior-based autonomy (or a ``behavior'') is defined as:
\begin{equation}
    \mathcal{U} = f_{behavior}(\bm{x}, r_1, r_2, \dots r_m),
    \label{eq:utility}
\end{equation}
where $\mathcal{U} \in {\rm I\!R}$ is the utility, $r_i \in \mathcal{S}_i$ is the $i^{th}$ decision variable in the corresponding domain $\mathcal{S}_i$, and $\bm{x}$ is the state of the agent and the perceived environment.

Oftentimes, an agent cannot take an action to maximize the utility of all its objectives simultaneously. A tradeoff is achieved through a scalarization according to

\begin{align}\label{eq:multi_obj_optimization_problem}
r_1^*, r_2^*, \dots r_m^* =& \argmax_{r_m \in \mathcal{S}_m \ \forall  m} \sum_{q=1}^{N_{beh.}} w_{q} f_{beh._q}(r_1, r_2, \dots r_m), 
\end{align}
where $N_{beh.}$ is the number of objectives, and each objective $f_{beh._q}$ is weighted by $w_{q}$. By solving for each decision variable $r_i$, an agent takes an action that maximizes the combined utility of all its objectives. 

\subsection{Safety Control Filters via Control Barrier Functions}

A nominal feedback controller for a USV with $p$ inputs, $\bm{u} = k(\bm{x}, r_1^*, r_2^*, \dots r_m^*) \in {\rm I\!R}^p$ determines the input $u$, such as thruster speeds, given vehicle state $\bm{x}$ and desired actions $r_1^*, r_2^*, \dots r_m^*$. 
The first order model of the closed loop dynamics of an agent's state is
\begin{equation}
    \dot{\bm{x}} = f(\bm{x}) + g(\bm{x})k(\bm{x}, r_1^*, r_2^*, \dots r_m^*).  \label{eq:cbf_dynamics}
\end{equation}

Control filtering is the process of restricting the nominal control inputs to a dynamically computed set of admissible values that ensure the vehicle will only operate in a safe state. 
From \cite{ames2017control}, the set consisting of all control values where the set $\mathcal{C}$ is forward invariant for the system \eqref{eq:cbf_dynamics} is defined as
\begin{equation}\label{eq:cbf_k_def}
    K_{cbf}(\bm{x}) = \{ u\in U : L_{f} h(\bm{x}) + L_{g} h (\bm{x}) u + \alpha ( h(\bm{x})) \geq 0 \}
\end{equation}
where details about the functions $\alpha$ and $h()$ can be found in \cite{ames2017control}. 
The safe set of trajectories, $\mathcal{C}$, can be derived from rigid safety requirements, such as avoiding imminent collisions.

\section{PROBLEM FORMULATION}

\subsection{Unifying Taxonomy}

\begin{table}[h!] 
\caption{Taxonomy of objectives for team mission}
\label{tab:objective_taxonomy}
\centering
\begin{tabular}{lll}
\hline
Objectives & Subtype & Description \\
\hline
Mission & Explicit & Directly specified team-level outcome \\
        & Implicit & Indirectly achieved through compliance \\
\hline
Behavioral & Coevolved & Optimized based on outcomes \\
           & Prescribed & Imposed by safety or regulatory norms \\
\hline
\end{tabular}
\end{table}

We introduce a taxonomy of mission and behavioral objectives in Table \ref{tab:objective_taxonomy}. 
Mission objectives are sparse team-level outcomes.
Explicit objectives are specified directly, while implicit objectives are achieved through compliance.

Behavioral objectives map actions to utility according to Eq. \ref{eq:utility}.
Coevolved objectives are optimized to support explicit mission outcomes, while prescribed objectives measure compliance with operational norms.
Each mission objective corresponds to a set of behavioral objectives: one per robot.

\subsection{Collaborative Swimmer Rescue Mission}
The motivating example used in this paper is a collaborative swimmer rescue mission, shown in Figure \ref{fig:real-world-overview}. This involves three high level mission objectives:

\begin{itemize}
    \item \textbf{Explicit Objective:} Maximize number of rescues
    \item \textbf{Implicit Objective:} Do not collide with any vehicles 
    \item \textbf{Implicit Objective:} Do not leave the operating region
\end{itemize}

\noindent
\textbf{Fitness Function:}
Our explicit mission objective that determines the number of swimmers rescued is formalized as

\begin{equation}
    G(T) = \sum_{j} \max_{t,i}(I(i,j)) \\
    \label{eq:Gsparse}
\end{equation}

where $T$ is the sequence of vehicle positions in a fixed time period. 
For each swimmer $j$, we check which vehicle $i$ was closest at any point in time $t$. 
The indicator function $I$ returns 1 if vehicle $i$ was within the capture radius of swimmer $j$ at time $t$ and 0 otherwise. 
If any vehicle $i$ captures swimmer $j$ at any point in time, we count that as a successful rescue.

\noindent
\textbf{Difference Fitness:}
We compute shaped fitnesses using

\begin{equation}
    D_i(T) = G(T) - G(T_{-i})
    \label{eq:difference_reward}
\end{equation}

where $T_{-i}$  counterfactually removes vehicle $i$'s positions from $T$. 
Intuitively, $G(T_{-i})$ determines how many swimmers would have been rescued without vehicle $i$, and $D_i(T)$ computes how many rescues vehicle $i$ is credited for \cite{agogino2004gecco}.

\subsection{Behavioral Objectives}

We use Eq. \ref{eq:utility} and \ref{eq:multi_obj_optimization_problem} from behavior-based autonomy to translate behavioral objectives into actionable decisions. Our vehicles make decisions about two variables: desired heading in the domain of 0-359 degrees and desired speed in the domain of 0-3 meters per second. 

Each mission objective has a corresponding set of behavioral objectives.
One set is ceovolved to achieve our explicitly defined team-level outcome of maximizing rescued swimmers. The other two are prescribed based on safety and regulatory norms. The prescribed behavioral objectives are a pair-wise COLREGS-based collision avoidance behavior between a vehicle and each nearby neighbor \cite{colregsbehavior}, and a Stay-In-Bounds behavior to return the agent to the operational region if they wander outside \cite{opregionbehavior}.
We show a vehicle-centric mapping of behavioral to mission objectives in Table \ref{tab:map}.

\begin{table}[h!]
\centering
\caption{Mapping behaviors to mission objectives}
\label{tab:map}
\begin{tabular}{ll}
\textbf{Behavioral Objective}                                                                               & \textbf{Mission Objective}                                                            \\ \hline
\begin{tabular}[c]{@{}l@{}}Derived: Take actions that\\ lead to more rescues\end{tabular}                   & \begin{tabular}[c]{@{}l@{}}Explicit: Maximize number\\ of rescues.\end{tabular}       \\ \hline
Prescribed: Adhere to COLREGs                                                                               & \begin{tabular}[c]{@{}l@{}}Implicit: Do not collide with \\ any vehicles\end{tabular} \\ \hline
\begin{tabular}[c]{@{}l@{}}Prescribed: Stay in bounds \\ if wandering away\end{tabular} & \begin{tabular}[c]{@{}l@{}}Implicit: Do not leave the \\ operating region\end{tabular}    \\ \hline
\end{tabular}
\end{table}

\subsection{Hardware Considerations}
We cannot accept collisions in our hardware experiments because this could damage our vehicles. Hence, we introduce a safety filter designed with control barrier functions \cite{ames2017control} as an additional layer of safety for hardware deployments.
The safe set $\mathcal{C}$ for a vehicle is defined as the subset of $x,y$ space on the plane that is 13 meters away from every other vehicle.

\begin{figure*}[ht!]
    \centering
    \includegraphics[width=0.9\linewidth]{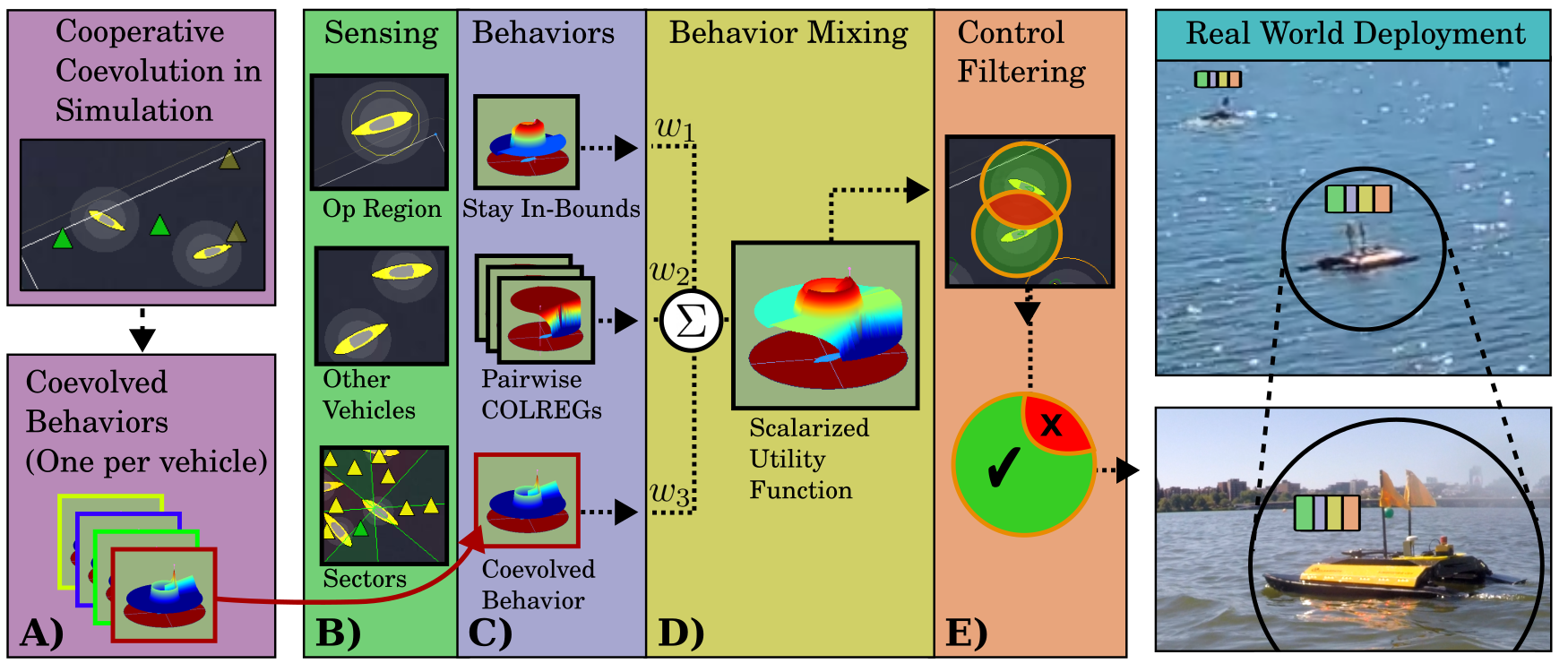}
    \caption{\textbf{Overview of MMOCIC.} \textbf{A)} Cooperative coevolution produces a set of coevolved behavioral objectives \textbf{B)} The vehicle aggregates input state information for each behavioral objective. \textbf{C)} Behavioral objectives produce utilities across the vehicle's action space. \textbf{D)} Scalarization produces a combined utility from behavioral objectives to determine the desired action. \textbf{E)} Control filtering prevents collisions in safety-critical deployments.}
    \label{fig:complete-overview}
    \vspace{-2mm}
\end{figure*}

\section{Method: Marine Multi-Objective Compliance-Integrated Coevolution}

\subsection{Architecture Overview}

A complete overview of Marine Multi-Objective Compliance Integrated Coevolution (MMOCIC) is shown in Figure \ref{fig:complete-overview}. MMOCIC follows a centralized training, decentralized execution paradigm. During training, cooperative coevolution is performed centrally in simulation, where teams of vehicles are jointly evaluated to produce one coevolved behavioral objective per vehicle. There is no central coordinator during simulated or hardware deployments. 

During a deployment, each robot runs its own decision-making stack. First, it aggregates the state information required to compute the utilities associated with its behavioral objectives (the (x) in Eq. \ref{eq:utility}). In the swimmer rescue mission, we define three behavioral objectives: (1) a coevolved rescue objective optimized through training, (2) a prescribed COLREGs objective \cite{colregsbehavior}, and (3) a prescribed Stay-In-Bounds objective \cite{opregionbehavior}. The Stay-In-Bounds objective activates when a vehicle approaches the edge of the operating region, while the pairwise COLREGs objective places high value on actions consistent with maritime right-of-way rules. The utilities produced by these objectives are combined through scalarization (Eq. \ref{eq:multi_obj_optimization_problem}), allowing each vehicle to balance rescue performance with compliant behaviors.

Although COLREGs promotes structured collision avoidance, it does not resolve in extremis situations where vehicles are already in close proximity 
\cite{international2003colreg,bakdi2022transportation,perera2018marine}. In our hardware deployments on a river, currents introduce unmodeled disturbances that can push vehicles directly into each other.
Since a collision could halt testing or damage our vehicles, we incorporate a control filtering layer based on Control Barrier Functions (CBFs), which can override the control input to prevent imminent collisions in extremis situations.

\subsection{Deriving Behaviors Through Cooperative Coevolution}

We train our vehicles using a Cooperative Coevolutionary Algorithm (CCEA) as described in \cite{gonzalez2025gecco}. We briefly summarize elements relevant to our contribution. Each vehicle evolves a neural network that maps sensor inputs to a control action. During deployment, we represent the neural network as a behavioral objective with high utility on the network's commanded action.
The CCEA maintains one population of neural networks per vehicle. From these populations, neural networks are sampled and assembled into teams and jointly evaluated using the sparse team fitness $G(T)$ and the corresponding difference fitness $D(T)$ (Eq. \ref{eq:Gsparse} and \ref{eq:difference_reward}). A selection mechanism determines which neural networks should be discarded and which ones should be kept for each generation's offspring. Selection is based on both team elitism and individual difference fitnesses, followed by mutation and re-evaluation over multiple generations.

Our swimmer rescue mission includes a stochastic element through randomized swimmer placement. To account for this variability, each assembled team is evaluated across multiple swimmer placement configurations, and fitness values are averaged prior to selection. This modification produces rescue behaviors that are robust to environmental variability rather than specialized to a single configuration.

\begin{figure*}[t]
    \centering
    \begin{subfigure}[]{0.32\textwidth}
            \includegraphics[width=\textwidth]{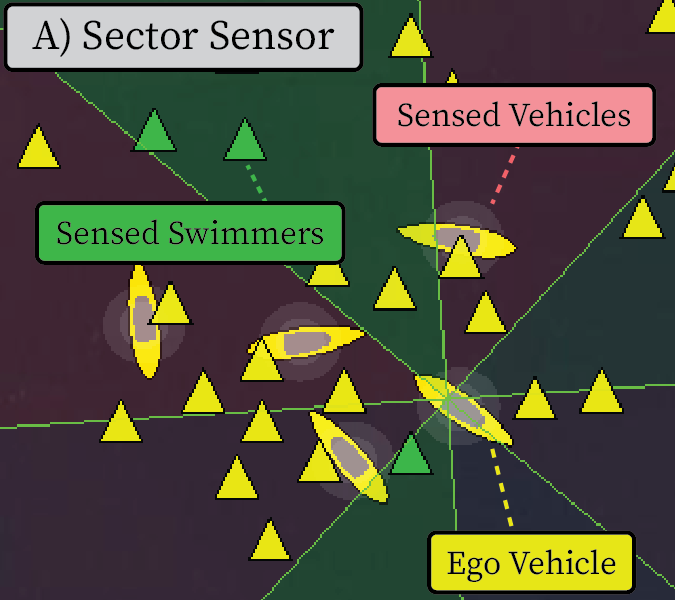}
            \label{fig:sectors} 
    \end{subfigure}
    \begin{subfigure}[]{0.32\textwidth}
            \includegraphics[width=\textwidth]{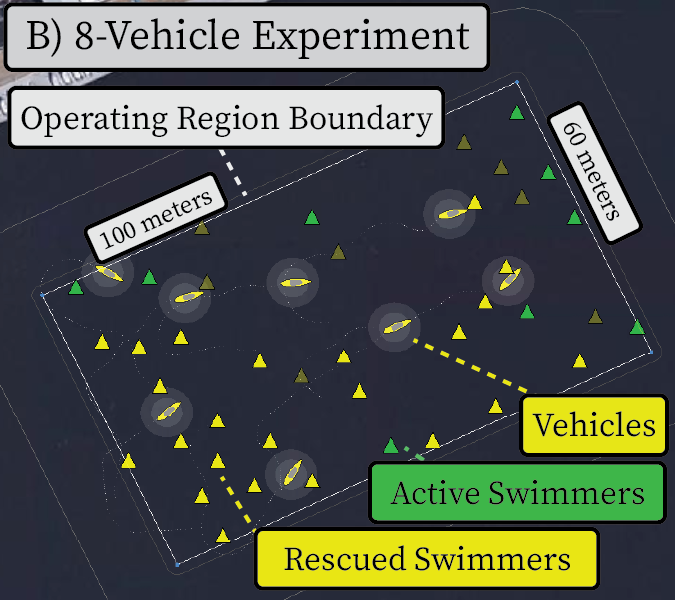}
            \label{fig:8v} 
    \end{subfigure}
    \begin{subfigure}[]{0.32\textwidth}
            \includegraphics[width=\textwidth]{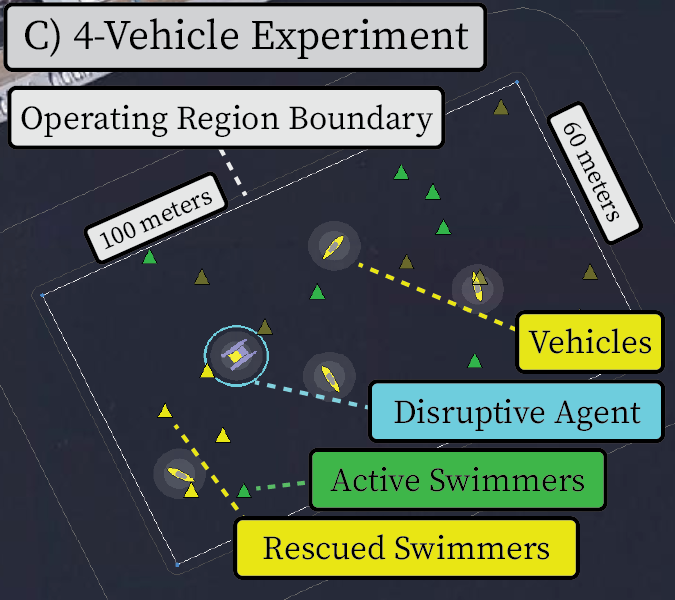}
            \label{fig:4v} 
    \end{subfigure}
    \vspace{-1mm}
    \caption{ \textbf{A) Sector sensor model.}  Green shading indicates swimmers are sensed in a sector. Red shading indicates other vehicles. Vehicles do not sense swimmers that have already been rescued. \textbf{B) 8-vehicle, 40 swimmer experiment.} 8 vehicles coordinate in a tight operating region to rescue 40 swimmers without collisions.
    \textbf{C) 4-vehicle, 20 swimmer experiment.} 4 vehicles coordinate to rescue 20 swimmers despite a disruptive agent that blocks them from rescuing swimmers.}
    \label{fig:triple}
    \vspace{-2mm}
\end{figure*}

\subsection{Balancing Behavioral Objectives}

During training, we can choose whether compliance-focused objectives are active or inactive. Keeping them active encourages the coevolved behaviors to negotiate with compliance behaviors for effective control, producing behaviors that optimize an explicit mission outcome given a fixed scalarization between behavioral objectives. This approach is appropriate when the balance between behavioral objectives is expected to remain unchanged between training and deployment. However, this commits the team to a single predefined tradeoff among high-level mission objectives.

Alternatively, compliance behaviors can be deactivated during training and reintroduced afterward. This allows coevolutionary optimization to explore a broader region of the behavior space without being shaped by a fixed compliance weighting. 
Once training is complete, compliance objectives can be reintroduced by increasing their weights, enabling dynamic exploration of tradeoffs among mission-level outcomes. This approach provides flexibility in settings where operational priorities may change, as it allows zero-shot adaptation to changing priorities without additional training.

\subsection{Control Filtering}

The COLREGS behavior gives a vehicle the ability to express its intention of maneuvering around other vessels, leaving execution details, such as adjusting the rudder, to a low-level controller. 
However, there are in extremis situations where desired control actions are not achievable given the vessel's dynamics \cite{international2003colreg, bakdi2022transportation, perera2018marine}. 
Furthermore, the collective utility function (see Eq. \ref{eq:multi_obj_optimization_problem}) may be weighted in a way that maximizes the coevolved behavioral objective, but the desired actions would lead to a collision. 
Thus, a CBF formulation \cite{usevitch2021acc} is implemented to maintain safety in hardware deployments, even when the intentions and actions of other vessels are unknown to the ego vehicle. ``Ego vehicle'' refers to the vehicle whose perspective we are considering. In the CBF formulation, the closed-loop dynamic model of each vehicle is assumed to be a single integrator unicycle model with a look-ahead linearization parameter $\gamma$. The advantage of this model is that there are no parameters in second-order dynamics to estimate, a difficult task to complete on the ego vehicle, but an impossible task to complete for other vehicles since their control inputs are not shared. Because of this limitation, each constraint in Eq. \ref{eq:cbf_k_def} assumes the other vehicle executes the worst case control input (i.e. turns hard towards the ego vehicle at full speed). 

\begin{table*}[]
\centering
\caption{\textbf{Metrics from 8-vehicle deployments.} Coevolved behaviors achieve comparable performance to the baseline with greater efficiency (lower total distance traveled).}
\label{tab:8-vehicle}
\begin{tabular}{lccccccc}
                   &           & Swimmers captured & Time to capture all & Total distance  & Close Encounters     & Near Misses          & Number of vehicles \\
Trial              & Strategy  & in 300 seconds    & 40 swimmers (s)     & traveled (m)    & (range \textless 8m) & (range \textless 4m) & out of bounds      \\ \hline
\multirow{2}{*}{1} & Baseline  & \textbf{40}       & \textbf{244}        & 2704.1          & \textbf{0}           & 1                    & \textbf{2}                   \\ \cline{2-8} 
                   & Coevolved & 38                & 475                 & \textbf{2402.6} & 5                    & \textbf{0}           & 4                   \\ \hline
\multirow{2}{*}{2} & Baseline  & \textbf{39}       & \textbf{390}        & 3647.3          & 5                    & 4                    & \textbf{2}                   \\ \cline{2-8} 
                   & Coevolved & 36                & 593                 & \textbf{2952.4} & \textbf{4}           & \textbf{0}           & 3                   \\ \hline
\multirow{2}{*}{3} & Baseline  & 36                & \textbf{346}        & 2320.8          & 3                    & 3                    & \textbf{2}                   \\ \cline{2-8} 
                   & Coevolved & \textbf{37}       & 442                 & \textbf{1592.6} & 3                    & \textbf{1}           & 4                   \\ \hline
\end{tabular}
\end{table*}

\begin{table*}[]
\centering
\caption{\textbf{Metrics from 4-vehicle deployments.} Efficiency encouraged by the difference fitness encourages coevolved vehicles to spread out, producing fewer or equal close encounters compared to the baseline.}
\label{tab:4-vehicle}
\begin{tabular}{lcccccccc}
                   &                      &           & Swimmers captured & Time to capture all & Total distance  & Close Encounters     & Near Misses          & Number of vehicles \\
Trial              & Adv.                 & Strategy  & in 300 seconds    & 20 swimmers (s)     & traveled (m)    & (range \textless 8m) & (range \textless 4m) & out of bounds      \\ \hline
\multirow{4}{*}{1} & \multirow{2}{*}{No}  & Baseline  & 17                & 357                 & 1417.9          & 3                    & 0                    & \textbf{0}                   \\ \cline{3-9} 
                   &                      & Coevolved & \textbf{19}       & \textbf{306}        & \textbf{1092.7} & \textbf{0}           & 0                    & 1                  \\ \cline{2-9} 
                   & \multirow{2}{*}{Yes} & Baseline  & \textbf{20}       & \textbf{225}        & \textbf{981.2}  & 3                    & \textbf{0}           & \textbf{0}                  \\ \cline{3-9} 
                   &                      & Coevolved & 18                & 340                 & 1262.2          & 3                    & 1                    & 1                   \\ \hline
\multirow{4}{*}{2} & \multirow{2}{*}{No}  & Baseline  & \textbf{20}       & \textbf{248}        & \textbf{1033.4} & 1                    & \textbf{0}           & 0                   \\ \cline{3-9} 
                   &                      & Coevolved & 19                & 307                 & 1167.0          & \textbf{0}           & 1                    & 0                   \\ \cline{2-9} 
                   & \multirow{2}{*}{Yes} & Baseline  & \textbf{19}       & 385                 & 1524.0          & 7                    & 0                    & 0                   \\ \cline{3-9} 
                   &                      & Coevolved & 17                & \textbf{370}        & \textbf{1488.5} & \textbf{3}           & 0                    & 0                   \\ \hline
\end{tabular}
\end{table*}

\section{IMPLEMENTATION DETAILS}

\subsection{Sensor Model}

We simulate ego-centric, low-resolution sensing for each vehicle's coevolved behavior based on true swimmer positions. 
This limited sensing reflects the fact that lost swimmers are visually difficult to detect in open water.
The sensor partitions the space around the ego vehicle into 8 sectors (Fig. \ref{fig:triple}A), producing 8 inputs for swimmers and 8 for vehicles.
The input from a sector is given by 

\begin{equation}
    S_{SEC.} = \frac{1}{(n_{TOTAL})(n_{SEC.})} \sum_{i'} \frac{(dist(i,i') - sat)}{(max - sat)}
    \label{eq:agentdensity}
\end{equation}

where $i$ represents the ego vehicle, and $i'$ iteratively represents entities sensed in this sector.
The maximum observation radius ($max$) is 100 meters. The saturation radius ($sat$) is 5 meters. 
$dist(i,i')$ measures the distance between $i$ and $i'$.
$n_{SEC}$ represents the number of entities detected in a single sector, while $n_{TOTAL}$ represents the total across all 8 sectors.

\subsection{Baseline Behavioral Objective}

We introduce a hand-designed behavior that can be substituted for a vehicle's coevolved behavior to provide an intuitive baseline for experiments.
This behavior aggregates sector inputs to determine the center of mass of all sensed swimmers. The relative bearing to the center of mass is given the greatest utility, and the speed is set to 1 meter per second.

\subsection{Parameters For Learning}

Each neural network is a multi-layer perceptron with tanh activation, 16 inputs, 10 units in the first hidden layer, 5 units in the second, and 2 units in the output layer that are each scaled, one for desired speed $\in [0,1]$, and relative heading $\in [-180,180]$. We include a mechanism that transforms relative $[-180,180]$ to an absolute $[0,359]$, as expected by Eq. \ref{eq:utility} for behavior mixing. The behavior's utility function has peaks at the desired speed and heading. 
Each population in the CCEA maintains 50 neural networks, and fitnesses are aggregated across 20 random swimmer configurations.

\section{EXPERIMENTAL SETUP}

We evaluated MMOCIC with deployments in hardware and simulation.
We set up experiments as three overarching missions. The operating region size across all missions is 100 meters x 60 meters. Missions are listed below.

\begin{enumerate}
    \item \textbf{Hardware:} 8-vehicle team, 40 swimmers, see Fig. \ref{fig:triple}B.
    \item \textbf{Hardware:} 4-vehicle team with disruptive agent, 20 swimmers, see Fig. \ref{fig:triple}C.
    \item \textbf{Simulation:} 4-vehicle team with 8 disruptive agents, 20 swimmers, similar to Fig. \ref{fig:triple}C.
\end{enumerate}

Our hardware deployments are risk-averse since collisions could halt further testing.
In simulation, we evaluated multi-objective decision support. This mission includes 12 vehicles in a tight operating region, 8 of which are disruptive agents that increase congestion. Since rescue vehicles are outnumbered 2 to 1, reaching a swimmer may require accepting compliance violations to maneuver through dense traffic.

\begin{figure}[t]
    \centering
    \begin{subfigure}[t]{0.49\linewidth}
        \centering
        \includegraphics[width=\linewidth]{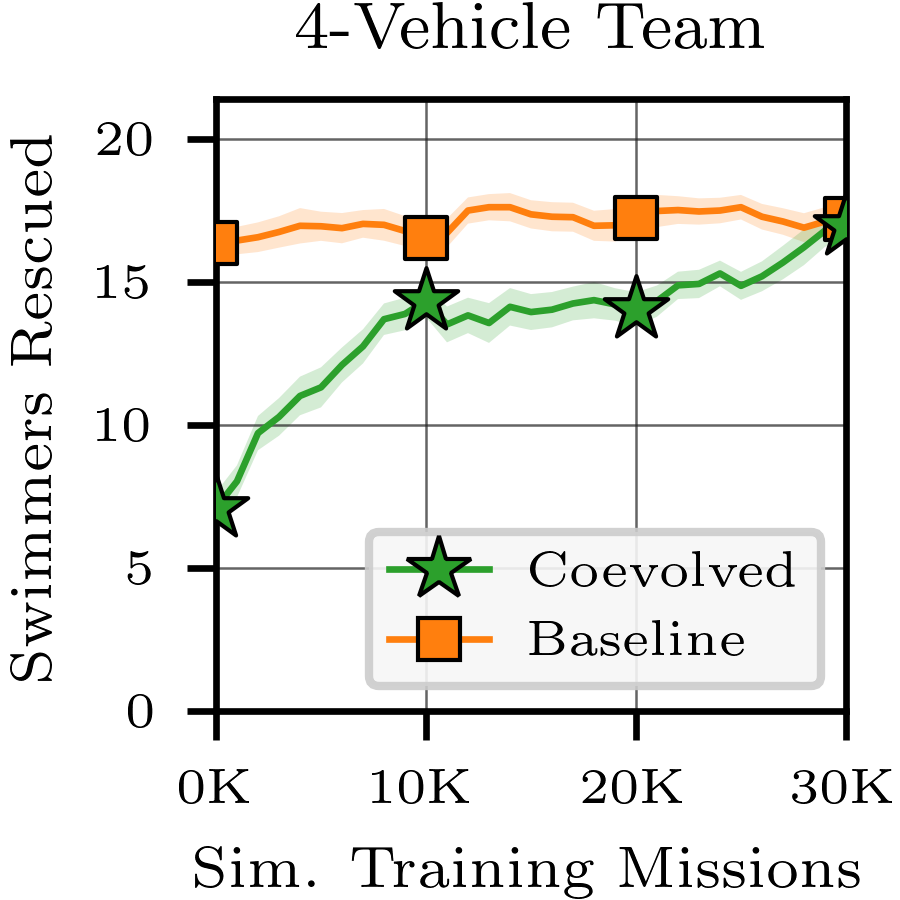}
    \end{subfigure}
    \hfill
    \begin{subfigure}[t]{0.49\linewidth}
        \centering
        \includegraphics[width=\linewidth]{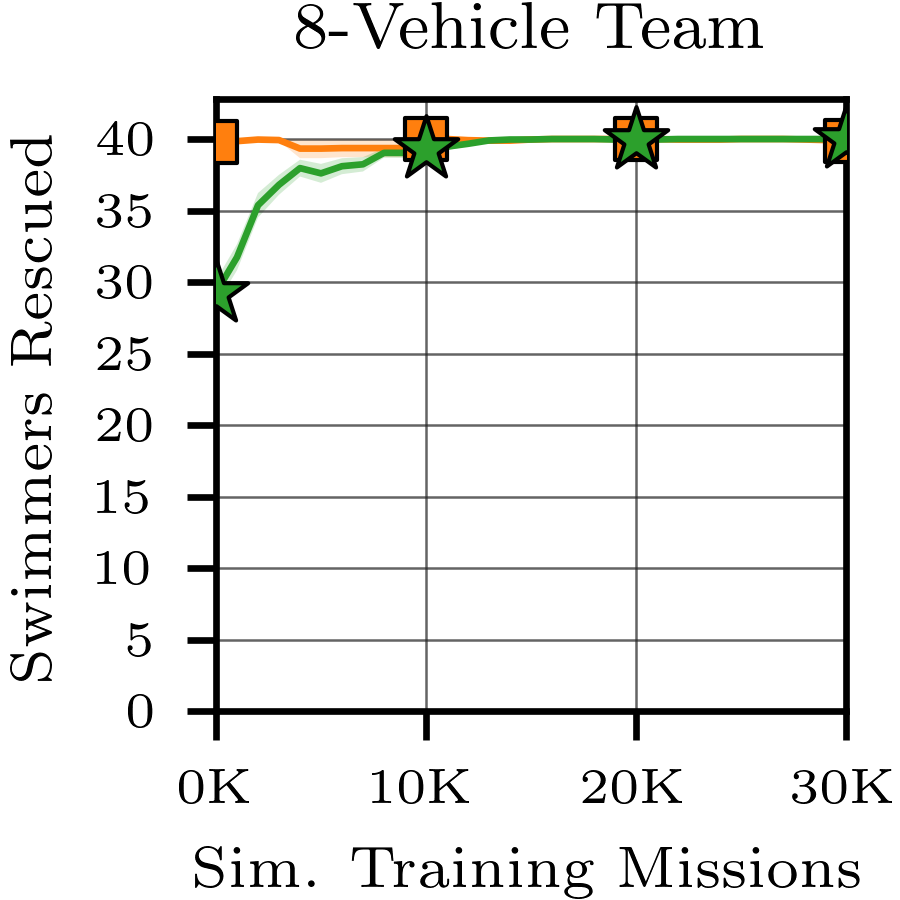}
    \end{subfigure}
    \caption{\textbf{Training Curves.} Coevolution produces collaborative rescue strategies through training that are competitive with a baseline behavior designed using domain expertise.}
    \label{fig:training}
\end{figure}

\begin{figure*}[ht!]
    \centering
    \begin{subfigure}[]{0.32\textwidth}
            \includegraphics[width=\textwidth]{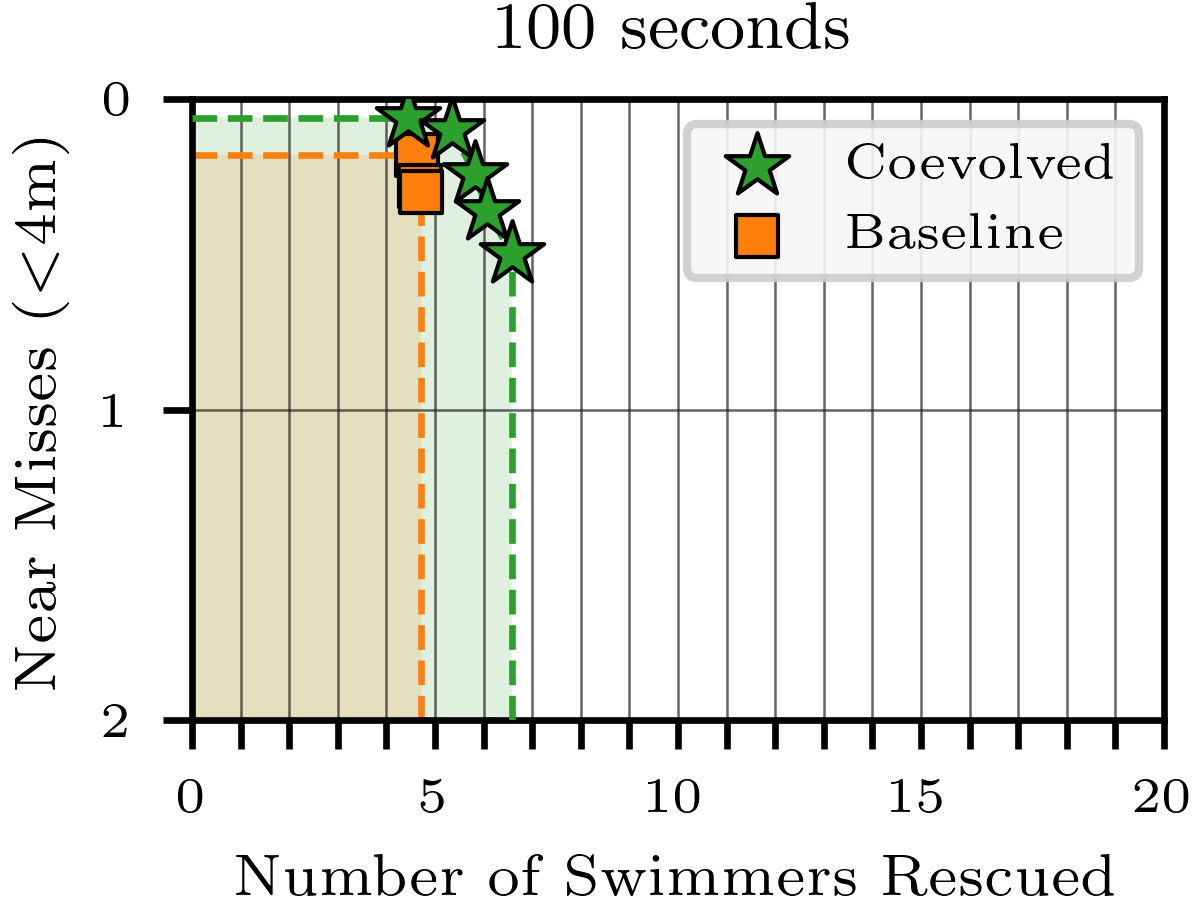}
    \end{subfigure}
    \begin{subfigure}[]{0.32\textwidth}
            \includegraphics[width=\textwidth]{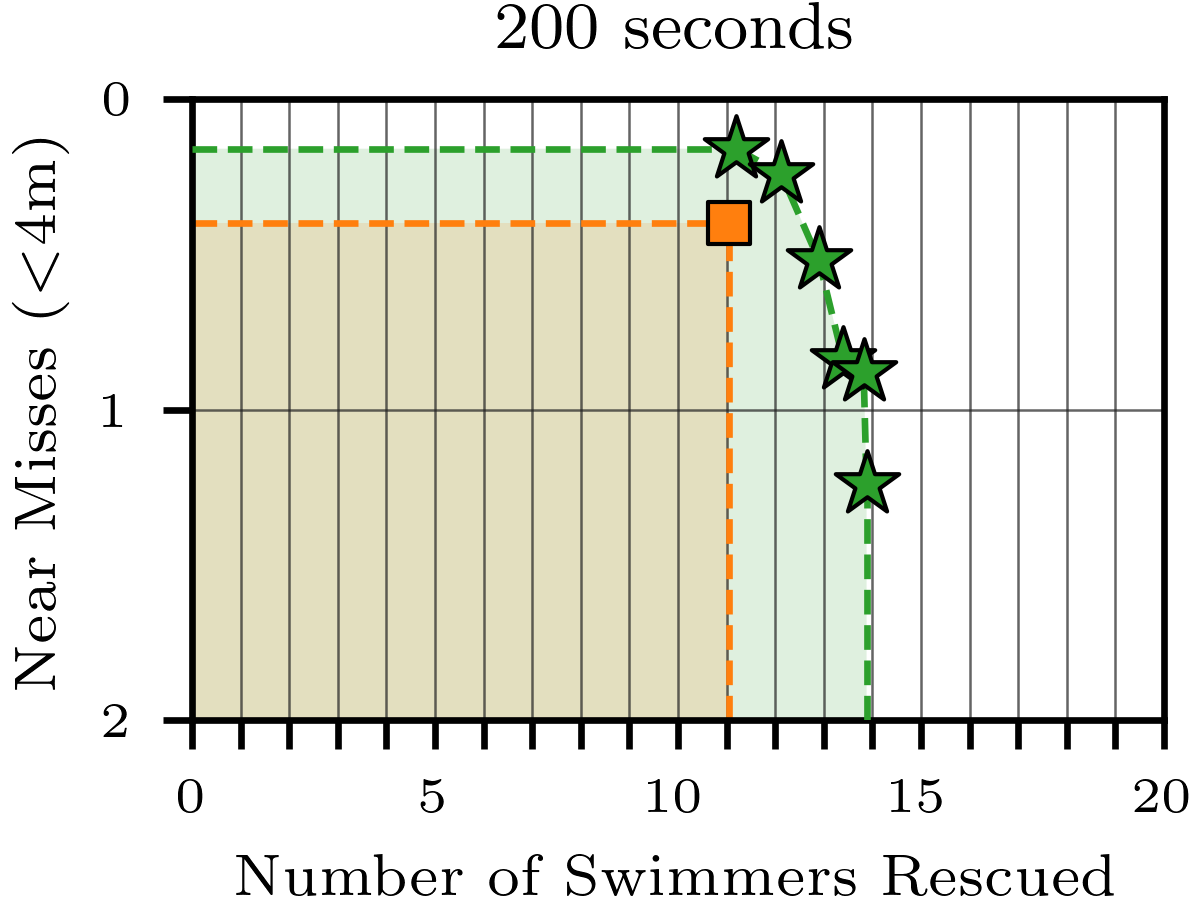}
    \end{subfigure}
    \begin{subfigure}[]{0.32\textwidth}
            \includegraphics[width=\textwidth]{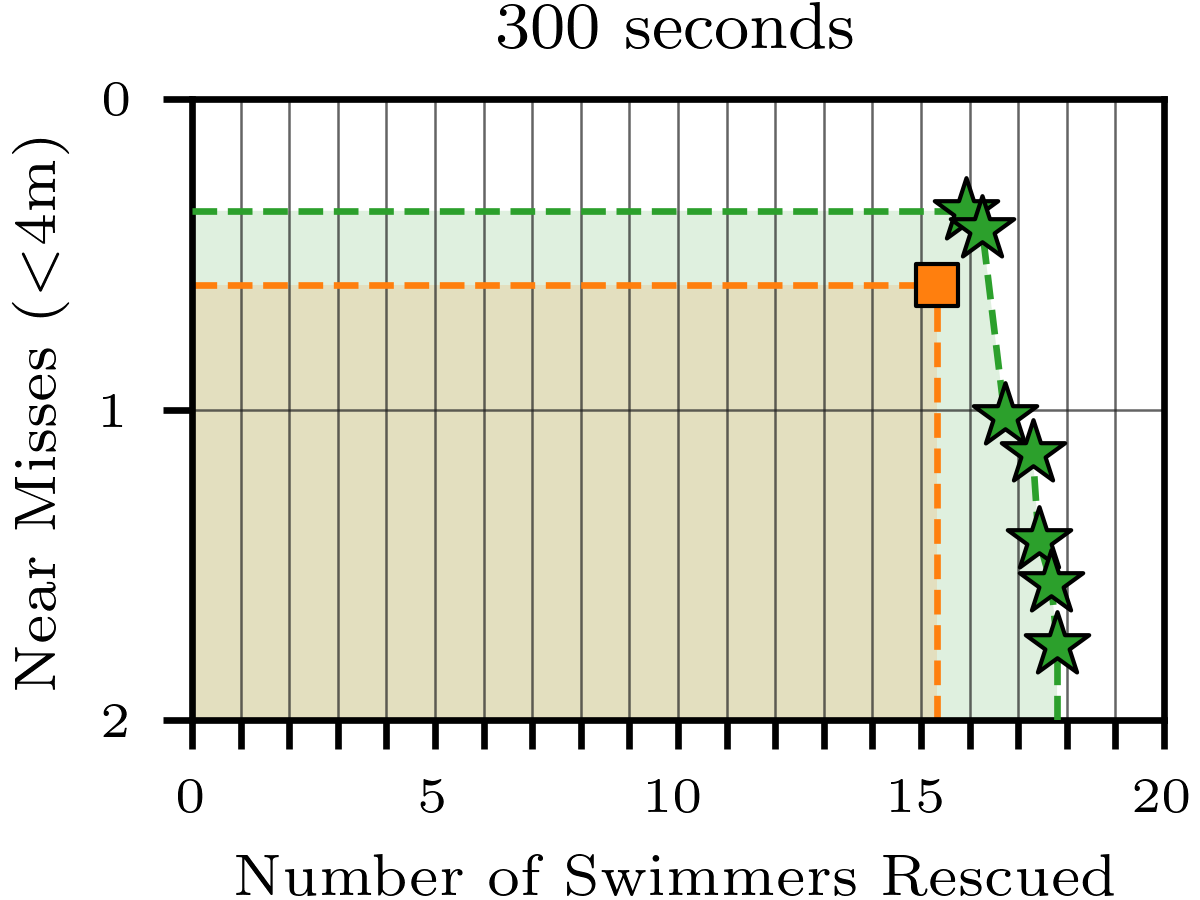}
    \end{subfigure}
    \vspace{-1mm}
    \caption{\textbf{Ablation Study.} Here we see how the best possible tradeoffs between different team level outcomes change throughout a simulated deployment. We see consistently better tradeoff options from the coevolved behaviors compared to the baseline. In fact, the coevolved behaviors always dominate the baseline, which is oftentimes reduced to a single solution.}
    \label{fig:pareto}
    \vspace{-2mm}
\end{figure*}

\subsection{Real-World Deployments}

We evaluated the performance of MMOCIC in three trials for the 8-vehicle mission, and two trials for the 4-vehicle mission. For each trial, we show performance and compliance metrics for the coevolved behaviors and the baseline behavior. Swimmer locations are randomized between trials but remain fixed within a given trial.  For added safety in the hardware experiments, we activated the CBFs to help prevent collisions in situations not covered by COLREGs.

In the 4-vehicle experiments, each trial is further split by whether we included the disruptive agent.
The disruptive agent was a human-operated vessel that purposely did not adhere to COLREGs. These experiments were designed to stress the system and evaluate how well the vehicles could respond to dangerous interactions that were not included in training.
The disruptive agent actively attempted to obstruct rescue vehicles from reaching swimmers and intentionally violated right-of-way rules to create hazardous situations. 

Before each hardware deployment, each team was trained with 30,000 simulated missions.  The scalarization between behavioral objectives was fixed throughout training and deployment. Training did not include any disruptive agents.

\subsection{Ablation Study In Simulation}

We trained a separate team distinct from those deployed in hardware so we could explore a tradeoff between performance and compliance. This team was coevolved over 100,000 simulated missions using a modified fitness function that awards a bonus proportional to remaining time if all swimmers are rescued before the 300-second limit. 

In principle, the CCEA can explore any behaviors regardless of additional behavioral objectives, but activating COLREGs in training partially reshapes any executed control commands. As a result, distinct rescue behaviors can be projected onto similar trajectories, obscuring meaningful performance differences.
The COLREGs behavior was deactivated during training so the coevolved behaviors had complete authority over the USVs to perform aggressive rescue maneuvers. There were no disruptive agents in training. 

After training, we performed a zero-shot ablation study by reintroducing the COLREGs behavioral objective without retraining. Disruptive agents were included in these tests and they navigated to random points in the operating region while adhering to COLREGs.
We adjusted the scalarization between COLREGs and the coevolved behaviors in 10\% steps to increasingly prioritize the COLREGs objective, systematically moving away from the original training balance of 100\% coevolved and 0\% COLREGs. For each fixed scalarization, we simulated 50 missions and recorded the average number of swimmers rescued and the average number of near-misses (when 2 or more vehicles are within 4 meters of each other).
This ablation provides a view of achievable tradeoffs between rescue performance and compliance. By selecting only nondominated solutions (maximizing rescues while minimizing near-misses), we obtain an approximate Pareto front over sparse team-level mission outcomes, even though the weighting is applied at the behavioral level.

\section{RESULTS AND DISCUSSION}

\subsection{Training for Real-World Deployment}

The training curves for the 4-vehicle and 8-vehicle teams are shown in Figure \ref{fig:training}, where swimmers rescued by the best team is plotted as a function of the number of simulated missions used for training. 
The number of swimmers rescued is reported as an average across 20 random swimmer configurations, with shading indicating standard error. 
For coevolved behaviors, we report the team achieving the highest average fitness within a generation. 
The baseline behavior is evaluated on the same configurations for direct comparison.

The CCEA for the 4-vehicle team starts with poor coordination, only rescuing approximately 7 of 20 swimmers. Through additional training, they rescue an average of 17 of 20 swimmers, which is comparable to the baseline behavior. 
The CCEA in the 8-vehicle study also starts with a lower performance than the baseline behavior, but quickly learns to rescue all 40 swimmers after just 10,000 simulations. 

\subsection{Real-World Deployments}

A quantitative summary is shown in Tables \ref{tab:8-vehicle} and \ref{tab:4-vehicle}.  Overall, the real-world deployments show a competitive performance between the coevolved teams and the baseline behavior teams. There is not a clear winner with respect to the number of swimmers captured in 300 seconds or the time to capture all swimmers. This reflects the results from training, where the coevolved behaviors produce competitive strategies compared to our baseline behavior.  

The safety and compliance metrics are similar for both the coevolved and baseline teams, suggesting that the COLREGs behavior, stay-in-bounds behavior, and CBFs were comparably effective when paired with either set of rescue behaviors.
One notable exception is that the coevolved teams traveled less total distance in the 8-vehicle trials, highlighting the impact of the difference fitness in Eq. \ref{eq:difference_reward}. If two vehicles visit the same swimmer, neither receives credit: removing either vehicle does not change the outcome because the other still completes the rescue. 
This discourages redundant effort and incentivizes teams to distribute coverage more efficiently.

In the 8-vehicle trials, a large portion of vehicles leave the operating region, occurring slightly more with the coevolved team. The rescue behavior can overpower the Stay-In-Bounds behavior to produce overly wide turns near the region’s edge, and could be addressed in future work by increasing the weight assigned to the Stay-In-Bounds behavior.

In the 4-vehicle trials, the coevolved team experienced fewer than or equal to the number of close encounters compared to the baseline, while near-miss counts and rescue metrics were comparable. The reduced number of close encounters is driven by the same efficiency observed in the 8-vehicle trials, where lower total distance traveled reflects more efficient coordination. Since the difference fitness encourages vehicles to avoid redundant coverage, they are less likely to cluster together, which in turn reduces the need to activate COLREGs or CBFs to avoid potential collisions.

\subsection{Ablation Study In Simulation}

Figure \ref{fig:pareto} presents the results of our ablation study, showing the progression of two approximate Pareto fronts over time. Each subfigure compares two Pareto fronts: one generated using coevolved behaviors and one using the baseline behavior, across two outcomes (number of swimmers rescued and near misses) at specific mission times (100s, 200s, or 300s). Each point represents a non-dominated solution averaged over 50 trials, with shading indicating which solutions are dominated by the Pareto front.

In this study we find that coevolved behaviors always dominate the baseline behavior, both in terms of avoiding near-misses and rescuing more swimmers. 
Moreover, these results indicate that in this mission with disruptive agents, it is possible to rescue more swimmers but only at the risk of additional collisions. 
These tradeoffs disappear using the baseline behaviors. In fact, a single point dominates all other points generated by the baseline behavior at 200s and 300s, leaving no meaningful choice between mission outcomes.

With the coevolved team, there is a benefit to adjusting the balance between behavioral objectives because they can achieve different mission-level outcomes. This presents an exciting opportunity for future work because weights may be adjusted online based on changing mission requirements.
For example, if a cautious rescue team that has been prioritizing collision avoidance is running out of time, it may shift its priority to accept a higher risk of collisions in order to save more swimmers before time runs out.

\section{CONCLUSION}

This paper presents Marine Multi-Objective Compliance-Integrated Coevolution (MMOCIC), a multi-robot framework that enables safe deployment of collaborative behaviors learned in simulation on real-world teams. 
By translating high-level mission objectives into low-level behavioral objectives, robots can make structured tradeoffs to balance mission goals with regulatory requirements. 
Future work should explore online adaptation mechanisms to adjust behavioral tradeoffs in response to changing mission conditions.

\addtolength{\textheight}{-21.3cm}   

\bibliographystyle{IEEEtran}
\bibliography{refs}

\end{document}